# Skip Training for Multi-Agent Reinforcement Learning Controller for Industrial Wave Energy Converters

Soumyendu Sarkar*[1], Vineet Gundecha[1], Sahand Ghorbanpour[1], Alexander Shmakov[1], Ashwin Ramesh Babu[1], Alexandre Pichard[2], Mathieu Cocho[2]

*Abstract*— Recent Wave Energy Converters (WEC) are equipped with multiple legs and generators to maximize energy generation. Traditional controllers have shown limitations to capture complex wave patterns and the controllers must efficiently maximize the energy capture. This paper introduces a Multi-Agent Reinforcement Learning controller (MARL), which outperforms the traditionally used spring damper controller. Our initial studies show that the complex nature of problems makes it hard for training to converge. Hence, we propose a novel "skip training" approach which enables the MARL training to overcome performance saturation and converge to more optimum controllers compared to default MARL training, boosting power generation. We also present another novel hybrid training initialization (STHTI) approach, where the individual agents of the MARL controllers can be initially trained against the baseline Spring Damper (SD) controller individually and then be trained one agent at a time or all together in future iterations to accelerate convergence. We achieved double-digit gains in energy efficiency over the baseline Spring Damper controller with the proposed MARL controllers using the Asynchronous Advantage Actor-Critic (A3C) algorithm.

## I. INTRODUCTION

Waves have been recognized as one of the efficient and consistent sources of renewable energy as we gradually move towards power generation with minimal carbon dioxide emission. Some of the already existing renewable energy sources that are widely deployed today are wind and solar [10][30]. Wave Energy can be valuable, specifically in the countries that have larger coastlines. Currently, the worldwide resource of coastal wave energy has been estimated to be over 2 Tera Watts which is 16% of the world energy consumption [17]. Hence, there is an increase in investment to build efficient Wave Energy Converters (WECs) to maximize the power produced. Some practical challenges that have been observed in deploying WECs in the ocean are the variable dynamics of the ocean waves, which make them vary in height, time period, and so on, especially in offshore locations. Hence, there has been ongoing research in developing larger and structurally robust systems to produce more consistent energy.

As part of this work, we focus on a state-of-the-art industrial three-legged multi-generator Wave Energy Converter system CETO 6 [1] as shown in Figure 1. Recent advancement in AI has enabled controllers in various domains to model and learn the complex representation of data and can efficiently model the movements of the structure and adjust themselves to stabilize [31][32]. Further, such controllers should also be able to react appropriately in case of scenarios that it has never seen before, which traditional learning techniques such as supervised learning [25] cannot solve. Hence, we proposed a novel reinforcement learning approach that consists of multiple agents controlling multiple legs of the WEC to learn the dynamics in our previous works [28]. These intelligent controllers not only help in stabilizing the structure but also in power maximization. As part of this work, we introduce a novel Skip training approach in a multi-agent setup that helps accelerate the convergence and recover the performance from "Falling off the cliff" during optimization. We show that our training approach achieves double-digit improvement in power generation compared to the baseline non-RL Spring Damper (SD) model. The major contribution of this paper can be summarized as:

1. A novel skip training approach has been introduced for a multi-agent RL which can effectively take a fully trained MARL out of performance saturation with incremental training(s) by a subset of RL agents and results in better performance.
2. The hybrid training initialization of RL agents with non-RL default controllers further facilitates skip training. This approach also simplifies and accelerates RL training convergence even without highly optimized MARL and as is explained later this is different from imitation learning.

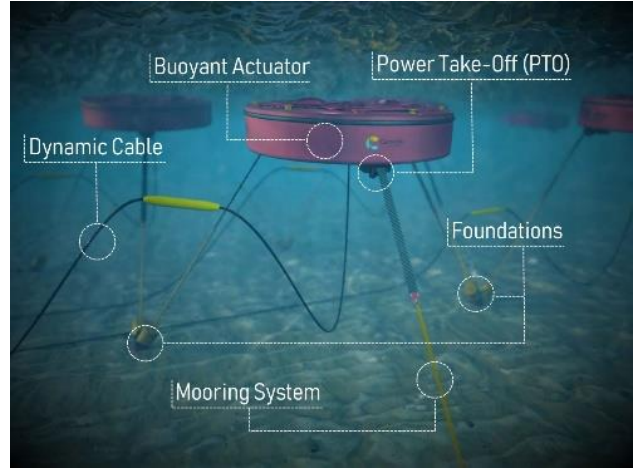

*Figure 1: CETO 6 Wave Energy Converter Platform* [1]

[1]Hewlett Packard Enterprise, USA  [2]Carnegie Clean Energy, Australia
{soumyendu.sarkar, vineet.gundecha, sahand.ghorbanpour, alexander.shmakov, ashwin.ramesh-babu}@hpe.com
{apichard, mcocho}@carnegie.com

The hybrid training initialization of RL agents with a non-RL default controller further facilitates skip training. This approach also simplifies and accelerates RL training convergence even without highly optimized MARL and as is explained later this is different from imitation learning.

## II. BACKGROUND AND RELATED WORK

### A. Structure of WEC

The state-of-the-art industrial WEC targeted in this study is shown in Figure 2. It has a cylindrical buoyant actuator (BA) that is submerged about two meters under the sea's surface. The buoy is secured to the seabed with three mooring legs, each of which terminates on one of the three power take-offs (PTOs) located within the BA [18]. The PTOs act like winches that can pull in (wind up) or let out (wind out) and adjust the tension on the mooring legs to vary in length. This converts 6 degrees of freedom translational and rotational motion of the BA at the point of tether connection into a linear motion. The electric generator of the PTO resists the extension of the mooring legs by applying varying reactive forces controlled by the RL controller, thereby generating electrical power. RL controllers need to optimize the timing and value of the PTO forces in relation to the wave excitation force, which is key to maximizing WEC energy capture and conversion efficiency.

### B. Mechanical Controllers for WEC

The Power Take-Off (PTOs) are composed of a mechanical spring and an electrical generator, as represented in Figure 2. There are multiple physical constraints to the PTO force and its components. The Wave energy source drives the reactive braking torque that acts against the input shaft. Hence the captured energy is equal to the braking mechanism of the system done by the generator minus the losses. The generator's spring component is modeled to induce resonance at the dominant wave frequency. This is analogous to impedance matching in the mechanical domain, where the impedance is effectively a measure of the opposition to motion when a potential force is applied. The average mechanical power generated by each PTO is the average of the product of the generator force and the retraction velocity. As this is the default controller of the WEC, we use this as a baseline to compare results with our proposed approach.

$$\overline{P_m} = \sum_{i=0}^{3} \overline{F_{gen_i} \times v_{PTO_i}}$$

$$F_{PTO} = Fspring + Fgen$$

Where, $F_{spring}$ is the spring force, and $F_{gen}$ is the generator force. $\overline{P_m}$ represents the power generated and $v_{PTO_i}$ represents the retraction velocity of the PTOs

### C. Reinforcement Learning for WECs

Reinforcement learning has been applied in the past for various control tasks in various applications[11][12][14][22]. They have achieved better performance than humans in applications like Atari [21] and game of go [26]. Reinforcement learning has also been popularly used for energy systems such as wind and power grids [15][16][17]

Specifically, for the Wave Energy Converter problem,

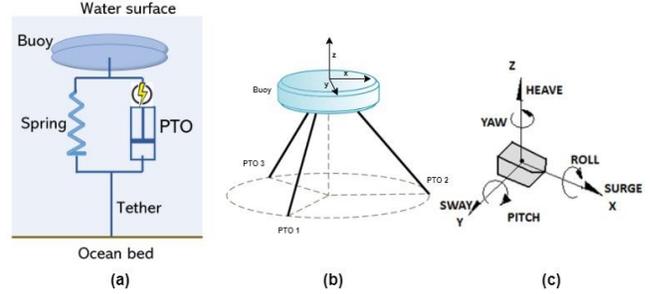

Figure 2: Geometry and parameters of the three-tether wave energy converter: (a) WEC spring, (b) 3D view, (c) PTO and motion with 6 degrees of freedom [24].

Reinforcement learning has been applied in the past but for a much simpler single-legged generator for both discrete and continuous action space [4][5][6]. Further, Anderlini et al. apply least-squares policy iteration for resistive control of a nonlinear model of a wave energy converter [2]. Work has been done using RL to obtain optimal reactive force for a two-body heaving point absorber with one degree of freedom (DOF) [2]. In the past, non-RL-based approaches have also been attempted to solve WEC optimization problems with Machine Learning and Deep Learning [3][32] but with certain limitations. However, all the previous approaches have been applied to one degree of freedom motion simple point absorber. **We were the first ones to use RL to control multi-legged WECs with six degrees of freedom of motion. Further, our novel skip connection approach to train a Multi-Agent RL controller has accelerated the training convergence significantly when compared to any other RL training strategies.**

## III. PROPOSED METHOD

### A. Problem Formulation

The multi-agent solution for the underlying problem is represented in Figure 3, the environment depicted on the left consists of the 3-legged wave energy convertor (WEC) and the wave sensors. The environment feeds the buoy's position, along with the velocity and acceleration, into the RL controller, along with values related to the leg extension and tension. It also feeds the oceanic waves of different heights and principal frequencies. Based on these inputs, the RL controller directs actions using the reactive forces on the three generators for the wave energy converter legs. The projected power generated in the three legs of the WEC is fed back to the RL controller as rewards. This feedback helps the RL controller assess the effects of its action based on inputs from the environment to take further actions based on the altered state.

From RL perspective, the system can be treated as,

$$S_{i+1} = f(S_i, A_i) \text{ where i} = 0,1,2\ldots\ldots$$

Where "i" is the time index. $S_i$ is the current state vector based on which an action vector "$A_i$" is applied for that time period which leads to a new state $S_{i+1}$. The action is subjected to the policy $\pi_\theta(S_i)$, which is parameterized by θ. Along with the state update, immediate reward "r" is calculated, which is represented as r($S_i$, $A_i$). The reinforcement learning problem can be solved by finding an optimum policy that maximizes the rewards' expected sum. Hence, the objective function can be defined as,

$$f(x) = max\ E_{s,A\ \sim \pi_\theta} \sum_{i=1}^{T}[\gamma^{i-1}\mathrm{r}(S_i, A_i)]$$

Where "T" represents the terminal state. s

### B. Environment States

For training, the environment state information is provided as a vector represented by s,

$$s = [\ e\ \dot{e}\ \ddot{e}\ g\ \dot{g}\ z\ \dot{z}\ ]^T$$

Where e represents the buoy position, g represents the tether extension, and z represents wave excitation. All RL agents share the continuous observation space of position and wave. The "dot" represents the first derivative (velocity), and "double dot" represents the second derivative (acceleration).

### C. Action

The actions of the MARL are the reactive forces on the three generators for the wave energy converter legs, which determine the resistance that the generator and PTO will offer to the movement of the buoy to generate electric energy.

The action space is continuous for the individual RL agent and is defined by the reactive force $f_{gen(k)}$ for the controlled generator, where "$k$" represents the index for the agent.

The target wave energy converter has limitations on the maximum and minimum tension in the spring extensions of the three legs anchored to the ocean bed to ensure mechanical integrity. Also, there are limits placed on the maximum reactive force on the generator, based on the generator ratings. In this design, as we explored the maximization of the energy capture, we made RL adapt to the hard limits by implementing WEC.

### D. Reward Shaping: Cooperation vs Competition

Specifying an appropriate reward function is fundamental to have the agent learn the desired behavior. Further, individual legs of the WEC consists of separate generators and the RL agents control either a single or a pair of legs, it is important to include the power contribution from all the generators in the reward. Although, it looks like a cooperative task from literature [19][23], the disparity in the in power generated by individual legs which is caused due to the trade-off by one or more legs to get additional power in other leg(s) makes the optimum solution a combination of co-operation and competition. So, a factor of flexibility is required to determine the extent to which power generated by the other legs are added to the reward of the individual legs. Further, an option for adversarial contributions of power from other legs are also required in the reward, which is achieved through, team coefficient, "η" where positive values add power generated by the other legs in the reward and vice versa.

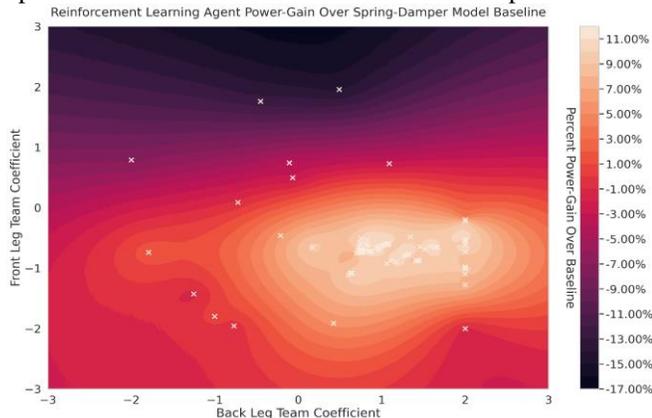

Figure 4: Team coefficient design for the front and back legs

$$Reward_k = P_{own(k)} + \eta_k \cdot P_{others}$$

Where P represents the generated power defined by, $-f_{gen}\ x\ \dot{e}$, η = team coefficient, $P_{own}$ is the power of the generator being controlled, and $P_{other}$ is power from other generators. We implemented a combined Bayesian hyperparameter search for the optimum 'team coefficient' of the individual agents for the back and front legs. We were able to converge to an optimum for the combined optimization, for 'team coefficient' of +0.8 for the agent for the back legs and -0.6 for the agent for the front leg, which can be observed from Figure 4.

### E. Reward Shaping: Cooperation vs Competition

In this work, we are using actor-critic based learning algorithms which have separate memory structure to explicitly represent the policy, independent of the value function. One neural network for computing an action based on a state and another neural network to produce the Q values (quality) for the action. The actor takes as input the state and outputs the best action. It essentially controls how the agent behaves based on the learning towards optimal policy, which is represented as, $\pi(a_t|s_t;\theta)$. The critic, on the other hand, evaluates the action by computing the value function represented as, $V = (s_t;\theta_v)$. The policy gradient method is used to optimize the policy (actor step) parallel with policy evaluation (critic step). Both policy and the value function are updated at the end of the terminal state. Hence, the objective function for policy gradients is defined as,

$$J(\theta) = \sum_{i=0}^{T-1} P(s_i, a_i|\tau) r_{i+1}$$

Where $r_{i+1}$ is the reward received by performing action $a_i$ at state $s_i$. Since this is a maximization problem, we optimize the policy by taking the gradient ascent with the partial derivative of the objective with respect to the policy parameter θ.

The update performed by the algorithm can be defined as,

$$\nabla_\theta J(\theta) = \sum_{i=0}^{T-1} \nabla_{\theta'} log\pi(a_i|s_i;\theta')\ A(s_i, a_i;\theta,\theta_v)$$

Where, $A(s_i, a_i;\theta,\theta_v)$ is an estimate of the advantage function, which is defined by,

$$A(s_i, a_i;\theta,\theta_v) = r_{i+1} + \gamma V_v(S_{i+1}) - V_v(s_i)$$

Where V represents the value function parameterized by "$v$."

Algorithm 1 represents the step-by-step details of the A3C [27] learning algorithm.

Simpler one agent RL with multiple actions failed to control the WEC effectively and was hard to converge to optimal solution and showed poor performance. Hence, separate agents of MARL were used to control the reactive force of the generators. A two-agent MARL is used for this

Algorithm 1: Reinforcement Learning Training

**T = 0, global parameters** $\theta$ and $\theta_v$
**Input**:
Environment state: buoy position (6 degrees of freedom), tether extension + preprocessing for 1st and 2nd derivates
Excitation: ocean wave episodes from JonSwap spectrum
**Reward Parameters:** team coefficient $\eta$

$$Reward_k = P_{own(k)} + \eta_k . P_{others}$$

**Initialization**: policy parameter $\theta_0$, value parameter
**Excitation parameters:** wave height, freq., direction
**Output**: Optimized Policy and value DNNs
**Repeat**
  $t_{start} = t$
  Sync. thread-specific parameters $\theta' = \theta$, $\theta_v' = \theta_v$
  Get initial state $s_0$
  **Repeat**
    Get $a_t$ according to the policy $\pi (a_t|s_t; \theta)$
    Get reward $r_t$, new state $s_{t+1}$
    t+=1
  **Until** $t = t_{max}$
  $R = V (s_t, \theta'_v)$ for non-terminal states,
  0 for terminal states
  **For** i in range (t-1, $t_{start}$):
    R ← $r_i + \gamma R$
    Accumulate gradients WRT $\theta'$
    Accumulate gradients WRT $\theta_v'$
  **End**
  Perform asynchronous update of $\theta$ and $\theta_v$
**Until** T >= Tmax

problem. The objective for using two agents was to exploit the symmetry of the two back legs, which were each 60 degrees apart from the axis of symmetry and duplicate the agent for the back legs while keeping a separate agent for the front leg, which is aligned to the axis of symmetry. This leads to optimal energy capture, with the challenge of convergence to better control policies. These challenges and solutions are described in the next section.

### F. Skip Training approach for RL convergence

Skip training shown in Figure 5 is our novel approach to solve the difficulty in converging MARL training to optimum solutions for complex systems like WEC. During the MARL training, once the peak has been achieved, it either saturates or falls off the cliff stalling the training from further improvement. Skip training introduces incremental training of a single agent while holding the other agents constant, thereby, creating a perturbation in the training trajectory to bring it back to a path towards better optima. Our results indicate a significant improvement of performance when skip training follows default MARL training.

The Hybrid training initialized (HTI) RL agents further assists skip training by enabling training of individual RL agents with default non-RL controllers active for rest of the system. In this method we first train individual RL controllers separately in hybrid mode, with non-RL state-of-the-art controllers like spring damper (SD) for the other leg(s), as seen in Figure 6(1). There are two cases. In the first case, we train RL for the front leg while the spring damper controls the back legs. For the second case, the back legs are trained by RL while the spring damper controls the front leg.

For cases where we would like to initialize all controllers, the RL agents goes through an alignment stage of multi-agent training. Else skip training can proceed on partially hybrid trained RL agents as described in Figure 6(2).

For further training we use skip training incrementally by training one agent at a time keeping the other agents fixed. This incremental training can proceed in a "Ping Pong" fashion till saturation is reached. It can also accommodate multi-agent incremental training. This architecture is shown in Figure 6(c). This approach simplified the training and yielded better results.

### G. Hybrid Initialization is Different from Imitation Learning

It may be noted that this approach is different from imitation learning. In imitation learning and behavioral cloning agents learn from expert system with feedback from demonstrations. On the contrary, our approach uses the expert as one of the actors, while the agent still learns from exploration and exploitation. This is done only during hybrid initialization. The reward from expert's action forms part of the total system reward. Further, the agents learn in a way to both cooperate and compete with the expert system for a better synchronization between the agents. Eventually, the non RL expert is replaced by another RL agent as the training progress forward and they start to explore. This process repeats, with one agent exploring and learning at a time.

## IV. EXPERIMENTS

In this section, we present the initial experiments to tune the hyper-parameters for A3C and select environment states that are most impactful and have positive contributions. As these were run with suboptimal RL agents, the power generation gains will show a lower value. For our skip training approach, we are limiting the number of agents to two for all our experiments.

### A. Hyperparameter tuning

We implemented a Bayesian search for optimized hyperparameters and often searched in pairs to keep the problem solvable. As most hyperparameters converged on dense islands in the hyperplane, as shown in Figure 7, it was feasible to thread them together to a combined set.

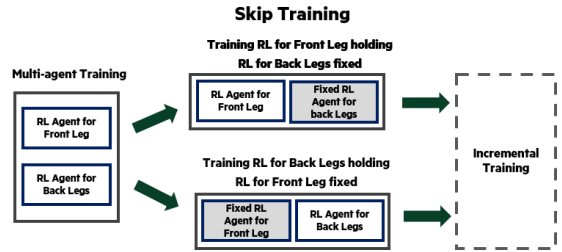

*Figure 5: Skip Training*

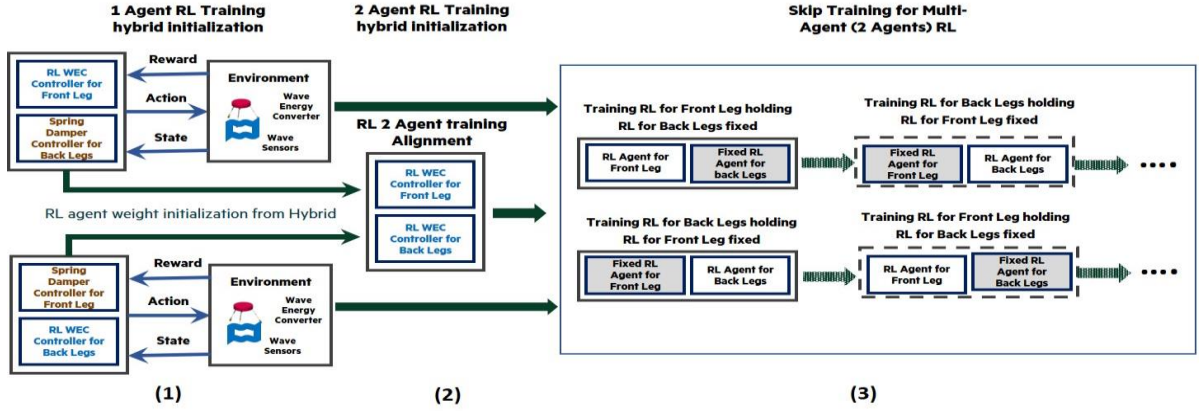

*Figure 6: Pipeline for Skip training with hybrid initialization. (1) Represents the individual agents for the front and back leg(s) trained against a non-RL Default Spring Damper (baseline) to enable skip training and learn from SD experience. (2) In cases where all RL agents are hybrid initialized, a multi-agent training is done for consolidation (3) In skip training phase, one agent is fixed while the other agent trains against the fixed agent(s) in a competitive way. This is done alternatively until the optimum performance is achieved. This stage can accommodate multi-agent training as well. The dotted box and arrow represent that they can be performed for any number of iterations.*

### B. Initial Experiment for Optimal Environment State Selection

We experimented with different sets of candidate environment states. We evaluated the effect of inclusion of these sets of states and retained only states whose inclusion showed improvements in total power gains and discarded the rest. We validated the states in successive steps and evaluated the impact of the choices for the two-agent RL controller architecture, which is randomly initialized and trained with the total power gain as a reward using A3C. We chose A3C as other algorithms such as SAC[7][13], PPO[20], A2C [9] and DQN [8] did not provide competitive results. This was done with untuned RL agents, which reflects low power gain improvements.

Step 1: Added position & velocity of buoy along different degrees of freedom along with wave elevation and rate of change of wave elevation.

Step 2: Added tether extension velocity (generator reactive forces are dependent on these)

Step 3: Added acceleration of the buoy

Step 4: Added tension value of the leg extensions (reactive forces must be clipped to limit tension)

It can be seen from Table 1 and Figure 8, that as the reactive generator forces depend on the leg extension velocities, the inclusion helped. The spring tension value in step 4 helped as the RL action of reactive force for generators needs to be clipped for physical limits, and this gives partial observability.

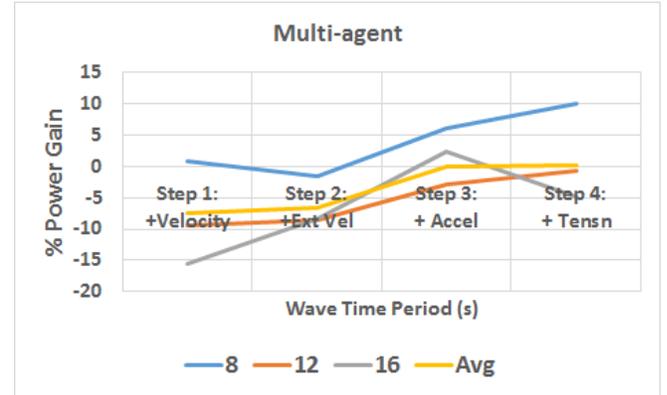

*Figure 8: %Power gain for each increment of environment state inclusion with initial unoptimized RL agent*

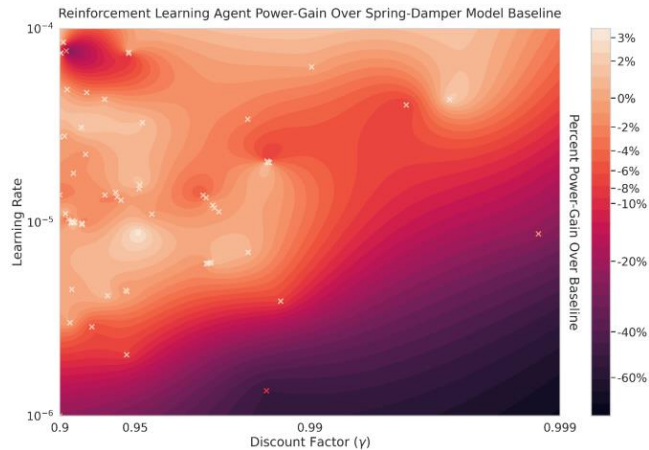

*Figure 7: Optimization for Learning rate and Discount Factor*

**TABLE 1: RL ENVIRONMENT STATE DESIGN (INITIAL UNOPTIMIZED RESULTS FOR STATE SELECTION)**

| Wave Period(s) | % Power Gain over Spring Damper as States are added to RL design ||||
| | Steps of State Design ||||
| | step 1 | step 2 | step 3 | step 4 |
|---|---|---|---|---|
| 8 | 0.9 | -1.6 | 6.1 | 10.1 |
| 10 | -5.3 | -5.0 | -2.7 | 0 |
| 12 | -9.4 | -8.6 | -3.0 | -0.7 |
| 14 | -11.1 | -11.7 | -1.9 | -1.9 |
| 16 | -15.5 | -8.3 | 2.3 | -2.8 |
| Avg | -8.08 | -7.04 | 0.16 | 0.94 |

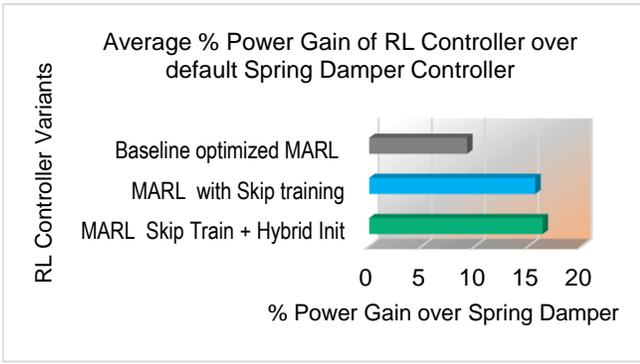

*Figure 10: % power gain comparison between conventional multi-Agent training(Hyperparameters tuned) vs Proposed Skip training.*

### C. Function Approximator Details

A fully connected neural network with two hidden layers of size 256 each was used as the function approximator. ReLU was used as the activation function between the layers and for the output layer.

## V. EVALUATION

The CETO 6 wave energy converter (WEC) platform simulator was used for this work, which accurately models the mechanical structure, the mechanical response, the electro-mechanical conversion efficiency with losses for generator and motor modes, and the fluid dynamical elements of the wave excitation.

Wave data such as the distribution of principal time periods, height, and spectrum were collected from Albany in Western Australia, Armintza in Spain (Biscay Marine Energy Platform: BiMEP), and Wave Hub on the north coast of Cornwall in the United Kingdom. The wave generator model used in simulation uses a well-established ocean wave spectrum like Jonswap which accurately models the heterogeneous components in ocean waves, letting the simulator sample the waves for training and evaluation. For evaluation, we used 1000 episodes for each principal wave period and height, where each episode covers 2000 sec of continuous-wave data in steps of 0.2 sec for the Reinforcement Learning (RL) loop and 0.05 sec (4x) for simulation response. For each training run, there are roughly 50 million steps for convergence, with 2 thousand such training runs required for hyper-parameter optimization and model search with early stops. For regular operation, we show results of median wave height of 2m for the entire wave frequency spectrum spanning time periods of 8s to 16s.

## VI. RESULTS

### A. Summary

The results indicate that reinforcement learning controllers have a double-digit % gain in power generation over the default non-RL Spring Damper (SD) controller, as shown in Table 2 and Figure 9. Our Skip-trained RL controllers performed better when compared to the hyperparameter optimized 2-agent RL controller trained with default random initialization. Also, the skip training with both the RL agents initialized with hybrid training performed best with an average of 16.3% gain over the default non-RL SD controller. For this RL agent, both the agents for the front leg and back legs are trained individually in hybrid mode with the SD controller followed by 2-agent RL training for alignment and subsequent skip training with the RL controller for the front leg incrementally trained while holding the RL controller for the back leg constant.

Also note that skip training was able to improve the power gain of the default MARL-trained RL controller from 9.2% to 15.6%, boosting the performance by 69% in this case. The intuition is that the skip training enables further optimization when the 2-agent RL has peaked.

**TABLE 2: RL Controller Power gain over default spring damper controller**

| Wave Time Period (s) | RL Agent Types | | |
|---|---|---|---|
| | Baseline optimized MARL | MARL with Skip training | MARL with Skip Training + Hybrid Initialization |
| 8 | 27.3 | 25.8 | 25.8 |
| 9 | 10.2 | 23.8 | 26.3 |
| 10 | 7 | 14.8 | 20 |
| 11 | 6.5 | 15.5 | 16.1 |
| 12 | 8.3 | 14.3 | 14.5 |
| 13 | 7.9 | 15.4 | 12.2 |
| 14 | 7.2 | 11.3 | 10.7 |
| 15 | 4.9 | 12.2 | 10.2 |
| 16 | 3.6 | 7.3 | 10.9 |
| **Avg** | **9.2** | **15.6** | **16.3** |

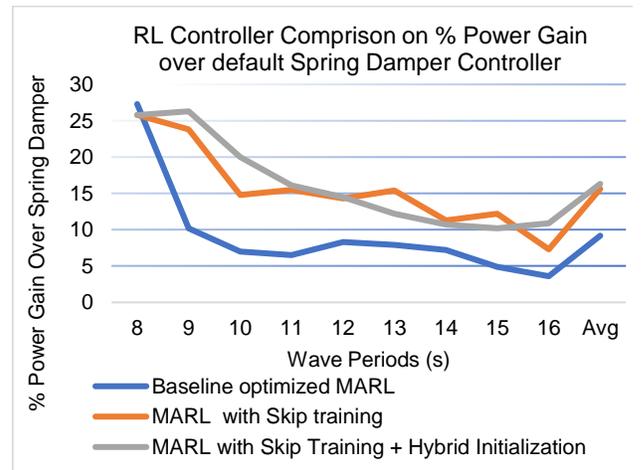

*Figure 9: %Power gain of different RL agents over default Spring Damper controller with variation of wave time period*

The single RL agent for the front leg trained in hybrid mode with SD controller followed by the RL agent of the back leg trained against the fixed initialized RL agent for the front leg (ping pong) yielded an average power gain of 10.7% over default SD controller, which is higher than the default MARL trained RL controller (9.2%). Note that this approach does not need the computationally expensive hyperparameter optimization like the MARL.

*B. Ablation Study*

Figure 10 shows the average % power gain of the multi-

-agent RL controller with different additive refinement of skip training and hybrid initialization.

For policy optimization algorithms we focused on Asynchronous Actor-Critic Agent (A3C) for this study. Through experiments we found that this algorithm performed better for the wave energy converter than other algorithms that we tried. DQN has limitations as it works best for discrete action space, and we needed to use continuous action space. As can be seen in Table 3, with Soft Actor-Critic we were unable to achieve better results.

**TABLE 3: Ablation study of RL algorithms:**
Percentage gain in energy capture of RL over Spring Damper. WTP: Wave Time Period

| WTP(s) | 8 | 10 | 12 | 14 | 16 | Avg |
|---|---|---|---|---|---|---|
| **SAC** | 13.7 | 0.5 | -1.4 | 1.1 | 3.2 | 3.2 |
| **A3C** | 25.8 | 20.0 | 14.5 | 10.7 | 10.9 | 16.1 |

## VII. CONCLUSION

The different architectures of the MARL controller described in this paper significantly improved the power generation over the non-RL default spring damper controller, evaluated over the entire spectrum of waves. This pushes wave energy converters closer to commercial success and helps in the decarbonization of energy production with clean energy.

However, a significant contribution of this paper is the **skip training** of multi-agent RL and the hybrid mode training of individual RL agents with default non-RL controllers for the rest of the system. Convergence to global optima is the biggest challenge for MARL agent architectures like A3C. In our study, we found that skip training with incremental training of one agent in a MARL, can bring the training out of saturation and help in convergence towards better optima in the complex hyperplane of MARL while improving performance by more than 60%. Also, the hybrid training initialization approach with skip training can achieve fast convergence without tedious hyperparameter optimization which often requires significantly higher computation and higher expertise. This applies to many other complex MARL control applications with multiple actors or entities to control. This includes robots navigating difficult terrains with a varying and unpredictable environment and other domain applications like wind power generators which also have similar problems where this technique can be effectively used to boost energy production.

For future work, we will expand our experiments to more than two agents with this skip training approach.